\documentclass[journal]{IEEEtran}

\usepackage[utf8]{inputenc} 
\usepackage[T1]{fontenc}    
\usepackage{hyperref}       
\usepackage{url}            
\usepackage{booktabs}       
\usepackage{amsfonts}       
\usepackage{nicefrac}       
\usepackage{microtype}      
\usepackage{graphicx}       
\usepackage{amsmath}        

\usepackage{amssymb}  
\usepackage{bm}  
\usepackage{textcomp}
\usepackage{enumerate}
\usepackage{multirow}
\usepackage{diagbox}
\usepackage{color}

\makeatletter  
\newif\if@restonecol  
\makeatother

\usepackage[linesnumbered,ruled,vlined]{algorithm2e}
\usepackage{algpseudocode}   

\begin{document}
\title{CHIP: Channel-wise Disentangled Interpretation of Deep Convolutional Neural Networks}

\author{Xinrui~Cui,
        Dan~Wang,
        and~Z.~Jane~Wang,~\IEEEmembership{Fellow,~IEEE}
\thanks{This work was supported in part by the Canadian Natural Sciences and Engineering Research Council (NSERC), the Four Year Doctoral Fellowship and the International Doctoral Fellowship at the University of British Columbia. (Corresponding author: Dan Wang.)

X. Cui, D. Wang, and Z. J. Wang are with the Department
of Electrical and Computer Engineering, University of British Columbia, BC, Canada. e-mail: (xinruic@ece.ubc.ca; danw@ece.ubc.ca; zjanew@ece.ubc.ca.)}}

\maketitle

\begin{abstract}
With the increasing popularity of deep convolutional neural networks (DCNNs), in addition to achieving high accuracy, it becomes increasingly important to explain how DCNNs make their decisions. In this work, we propose a CHannel-wise disentangled InterPretation (CHIP) model for visual interpretations of DCNN predictions. 
The proposed model distills the class-discriminative importance of channels in DCNN by utilizing sparse regularization.
We first introduce network perturbation to learn the CHIP model.
The proposed model is capable to not only distill the global perspective knowledge from networks but also present class-discriminative visual interpretations for the predictions of networks.
It is noteworthy that the CHIP model is able to interpret different layers of networks without re-training.
By combining the distilled interpretation knowledge at different layers, we further propose the Refined CHIP visual interpretation that is both high-resolution and class-discriminative.
Based on qualitative and quantitative experiments on different datasets and networks, the proposed model provides promising visual interpretations for network predictions in image classification task compared with existing visual interpretation methods.
The proposed model also outperforms related approaches in the ILSVRC 2015 weakly-supervised localization task.
\end{abstract}

\begin{IEEEkeywords}
Model interpretability, visual interpretation, channel-wise disentanglement, network perturbation.
\end{IEEEkeywords}

\section{Introduction}
\label{sec:intro}
Deep convolutional neural networks (DCNNs) have achieved impressive performances in many computer vision tasks such as image classification \cite{huang2017densely}, object detection \cite{pang2019libra}, image captioning \cite{anderson2018bottom}, and visual question answering \cite{wang2018fvqa}. However, the complex internal working mechanism of DCNNs remains elusive and makes them highly non-transparent to human \cite{Lipton:2018:MMI:3236386.3241340}. The lack of interpretability of DCNNs may result in a general mistrust of their results and further hinder their adoption, especially for critical applications such as medical diagnosis and criminal justice \cite{8099861}. 

Interpretability study has drawn increasing attention, e.g., for building trust for real-world users and offering insight into black-box models for practitioners \cite{rahwan2019machine, AAAI1714434}. 
Visual interpretation is one prevalent interpretation method. 
A number of visual interpretation models were proposed to interpret the decisions of DCNNs \cite{8237633, zhou2016learning, 8237336}. A desired visual interpretation model should possess the following properties:
\begin{enumerate}[$\bullet$]
\item The model should give class-discriminative visual interpretation, because the DCNN being explained is designed to distinguish different classes.
The class-discriminative visual interpretation should highlight the critical region of the input for a specific class prediction.
\item The model should be able to present the fine-grained information that is important to the class prediction.
\item The model is supposed to obtain interpretable knowledge from different layers and lend insight into the roles of different layers of DCNNs.
\item Given an image, the model should give a specific explanation for the network decision.
People can decide whether to trust that decision based on the instance interpretation. 
\item The model should give global interpretation of networks. General interpretation knowledge should be distilled from a large image dataset rather than from a single image.
\end{enumerate}

It should also be noted that there is a trade-off between the performance and the interpretability of DCNNs. 
Pursuing the interpretability of DCNNs may sacrifice their accuracies, which is undesirable. We therefore focus on the post-hoc interpretability which refers to the extraction and analysis of information from a trained network.
In this situation, the network usually does not have to sacrifice its performance in order to be interpretable.

For visual interpretation, several approaches have been proposed to highlight a portion of internal features that are critical to the decision by ascribing saliency \cite{zhou2016learning, 8237336, springenberg2014striving, zeiler2014visualizing}.
However, these methods have certain limitations.
For example, Guided Backpropagation \cite{springenberg2014striving} cannot give class-discriminative visualization.
Class Activation Mapping (CAM) \cite{zhou2016learning} and Grad-CAM \cite{8237336} are only constrained in the last convolutional layer to present the visual interpretation.
Table \ref{cmp_table1} shows the properties of different visual interpretation methods.

\begin{table*}[]
\caption{Properties of different visual interpretation methods}
  \label{cmp_table1}
  \centering
\begin{tabular}{|c|c|c|c|c|c|c|c|}
\hline

\multirow{2}{*}{\backslashbox{Method}{Property}} & \multicolumn{2}{c|}{Instance interpretation} & {General} & Class-discriminative  & High-resolution  &Multi-layer & {Post-hoc } \\ \cline{2-3}
&Perturbation-based &Gradient-based  &interpretation  &interpretation  &interpretation  &interpretation &interpretation\\ \hline
                  
Deconvolution\cite{zeiler2014visualizing}&   &\checkmark &    &     &\checkmark &\checkmark &\checkmark  \\ \hline

Guided & \multirow{2}{*}{} & \multirow{2}{*}{\checkmark} & \multirow{2}{*}{} & \multirow{2}{*}{} & \multirow{2}{*}{\checkmark} & & \multirow{2}{*}{\checkmark} \\
Backpropagation\cite{springenberg2014striving} &   &  &    &     &     & & \\ \hline

CAM\cite{zhou2016learning}&  &\checkmark &    &\checkmark  &     &     &    \\ \hline

Grad-CAM\cite{8237336}&    &\checkmark &     &\checkmark  &   &    &\checkmark  \\ \hline

Guided & \multirow{2}{*}{} & \multirow{2}{*}{\checkmark} & \multirow{2}{*}{} & \multirow{2}{*}{\checkmark} & \multirow{2}{*}{\checkmark} & & \multirow{2}{*}{\checkmark} \\
Grad-CAM\cite{8237336}&    &    &    &     &      &   &     \\ \hline

LIME\cite{Ribeiro2016}&\checkmark &   &    &\checkmark  &    &  &\checkmark  \\ \hline

DeepDraw\cite{deepdraw}&     &   & \checkmark  & \checkmark  & \checkmark &  & \checkmark  \\ \hline

Guided CHIP&\checkmark &\checkmark & \checkmark  &\checkmark  &\checkmark  &\checkmark &\checkmark \\ \hline

CHIP \& & \multirow{2}{*}{\checkmark} & \multirow{2}{*}{} & \multirow{2}{*}{\checkmark} & \multirow{2}{*}{\checkmark} & \multirow{2}{*}{\checkmark} & \multirow{2}{*}{\checkmark} & \multirow{2}{*}{\checkmark}\\
Refined CHIP& &  &    &      &   &    &  \\ \hline
\end{tabular}
\end{table*}

In order to alleviate the above issues, we propose a CHannel-wise disentangled InterPretation (CHIP) model which can build trust in networks without sacrificing their accuracies.
The main contributions of this work are summarized as follows:

\begin{enumerate}[$\bullet$]
\item The proposed CHIP model can disentangle channels in a network to interpret internal features in different layers.
Based on that, CHIP can give visual interpretations to network decisions without re-training.
\item In CHIP, we first introduce the channel-wise network perturbation method by perturbing network features.
The underlying principle is that the class prediction would drop dramatically if the forward propagation of class-specific important channels is blocked.
Our model also utilizes the inherent sparse property of net-features as a regularization.
\item CHIP can distill the class-discriminative knowledge of a network from a large dataset and utilize the distilled knowledge to interpret the net decision for an instance.
It means our model can not only explore the general interpretation information without data bias but also give a visual interpretation for the prediction of a particular input.
\item Through combining CHIP visual interpretations in the shallow and the deep layers, we extend CHIP to Refined CHIP which can give a high-resolution class-discriminative interpretation.
\item The proposed CHIP model can be applied to tackle the weakly-supervised localization task. It achieves better performance on ILSVRC 2015 dataset when compared with previous related work.
\end{enumerate}

\section{Related Work} \label{relate}
The most related studies to our work are visual interpretation of DCNNs and its application in the weakly-supervised localization, as described below:

\textbf{Visual interpretation of DCNNs.} 
Visual interpretation of DCNNs can be classified into two major categories: one is instance interpretation which means the interpretation for a particular input; and the other is general interpretation which denotes the interpretation of the overall network.
Both focus on providing visual interpretation to explore the internal working mechanism of DCNNs. Table \ref{cmp_table1} shows different visual interpretation methods and their properties.

Methods in the first category can be further divided into gradient-based and perturbation-based.
Gradient-based approaches give interpretation by using the gradient (or gradient variant) of the output or the internal unit.
For instance, Guided Backpropagation \cite{springenberg2014striving} and Deconvolution \cite{zeiler2014visualizing} interpret the network by visualizing internal units.
They design different backward pass ways to map neuron activation down to the input space, visualizing the input image pattern that is most discriminative to the neuron activation.
Although they can give high-resolution visualization for a specific input image, the visualization is not class-discriminative.
In contrast, CAM \cite{zhou2016learning} and Grad-CAM \cite{8237336} are able to give the class-discriminative interpretation. 
Specifically, they visualize the linear combination of activations and class-specific weights in the last convolutional layer.
However the obtained visualization is not high-resolution and does not show fine-grained details.
To tackle this problem, Guided Grad-CAM was designed by combining Guided Backpropagation visualization.
Another common drawback is that their visual interpretation is only effective for the last convolutional layer.
In addition, CAM can only provide interpretation for a specific kind of DCNN architecture.

Perturbation-based methods mainly focus on the input image perturbation \cite{8237633, Ribeiro2016}. 
To be specific, they involve perturbing input images and observing the change of predictions. 
The intuitive reason is that the prediction would drop by the maximum amount when pixels contribute maximally to the prediction are modified.
Although these methods can obtain the class-discriminative visualization, they do not explore the internal mechanism of networks or interpret internal features. 
Meanwhile, their performances are affected by the size and the shape of occluded pieces in perturbed images.
For instance, image patches in regular grids are used for occlusion in \cite{zeiler2014visualizing} while super-pixels from segmentation are used in \cite{Ribeiro2016}.

Methods in the second category focus on visualizing DCNNs in a global perspective by synthesizing the optimal image that gives maximum activation of a unit \cite{deepdraw, olah2017feature}. 
Although these methods are able to give class-discriminative interpretation, they are not designed for interpreting specific input images.
In some sense, they only provide high-level abstract interpretation and are not helpful to understand specific decisions of DCNNs.

There is a compromise between instance interpretation and general interpretation: 
Instance interpretation can explain network predictions for a given input; while the knowledge obtained by the general interpretation is more stable and representative without data bias.
To balance the desired characteristics of two interpretation categories, our proposed interpretation model is designed to distill the class-discriminative knowledge of a network from a large dataset and then utilize the knowledge to explain specific predictions.
Further, Refined CHIP is designed to provide visual interpretation that is both class-discriminative and high-resolution.

\textbf{Weakly-supervised Localization.}
This task refers to localizing the object only with the image-level class label. One approach is to utilize the visual interpretation distilled from the network to localize the target object \cite{zhou2016learning}, \cite{8237336}.
From this point of view, visual interpretation models can be applied to weakly-supervised localization.

In \cite{zhou2016learning}, a network should be modified to a particular kind of architecture to learn class activation maps which can be used to generate predicted bounding boxes for weakly-supervised localization.
In the modified architecture, the convolutional layer is followed by the global average pooling layer and then the softmax layer.
Compared with the original network, the modified network architecture may achieve inferior classification accuracy.
Therefore, the localization accuracy would also be limited by the inferior accuracy of the modified network.
In Grad-CAM, the gradients of the output to feature maps are used as the weights of feature maps to obtain class activation maps.
Grad-CAM does not need to modify the network architecture and can be applied to different networks.
It should be noted that the class-discriminative weights in Grad-CAM are only obtained from a single image.
In contrast, our proposed CHIP model distills the class-discriminative knowledge from a large dataset, which provides more insight into the network and can also enhance the localization accuracy.

\section{Methodology}
\label{sec:methodology}
In this section, we present the proposed CHIP model whose objective is to provide insight into DCNNs based on the distilled class-discriminative knowledge.
In section \ref{3a}, we describe the disentangled channels based on perturbed networks. 
In section \ref{3b}, we present the CHIP model to distill important channels for different classes. With this model, we also describe the way to obtain class-discriminative visual interpretations for the decisions of networks and apply it to weakly-supervised localization. Section \ref{3c} describes the proposed CHIP algorithm.

\subsection{Disentangled channels based on Perturbed Networks}\label{3a}
$\textbf{Class-discriminative importance of channels.}$ 
To disentangle the roles of channels, our interpretation model is designed to learn the importance of channels for different classes.
Through turning off the forward propagation of partial channels each time, we can learn the class-discriminative importance of channels by analyzing the variation of predictions.
The underlying principle is that the class prediction would drop dramatically if the forward propagation of important channels is blocked.

Specifically, for the $l$-th layer, the class-discriminative importance matrix is denoted as $\mathbf{W} = [\mathbf{w}_1 ~ \mathbf{w}_2 \cdots \mathbf{w}_c \cdots \mathbf{w}_C]^T$, where $C$ is the number of image category.
For the $c$-th class, the class-discriminative importance is $\mathbf{w}_c = [w_c^1 ~ w_c^2 \cdots w_c^k \cdots w_c^{K}]$, where $w_c^k$ represents the importance of the $k$-th channel and $K$ is the number of channels in the $l$-th layer.

\begin{figure}[htp]
	\centering
	\includegraphics[width=1\linewidth]{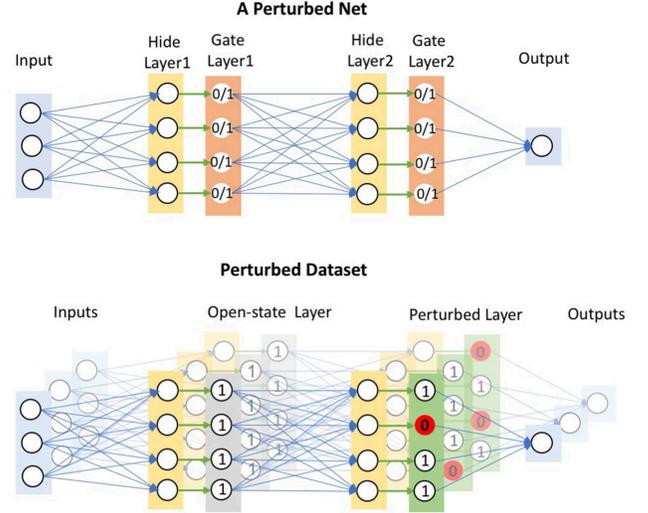}
	\centering
	\caption{Illustration of a perturbed network and the perturbed dataset.}
	\label{fig:illustration1}
\end{figure}

$\textbf{Channel-wise perturbed networks.}$ 
Inspired by the perturbation-based method \cite{Ribeiro2016}, we perturb a pre-trained network $f(\cdot)$ by channel gates to learn the class-discriminative importance of channels. 
As shown in Fig. \ref{fig:illustration1}(top), each original layer is associated with a gate layer in which each channel gate controls the state of the corresponding channel in the former layer.
Here, in the $l$-th layer, we use a binary vector $\mathbf{d} = [d_1 ~ d_2 \cdots d_k \cdots d_{K}]^T$ to denote the channel gate layer.
The $k$-th channel is turned off if $d_k$ is zero.
\begin{figure*}[htp]
	\centering
	\includegraphics[width=0.6\linewidth]{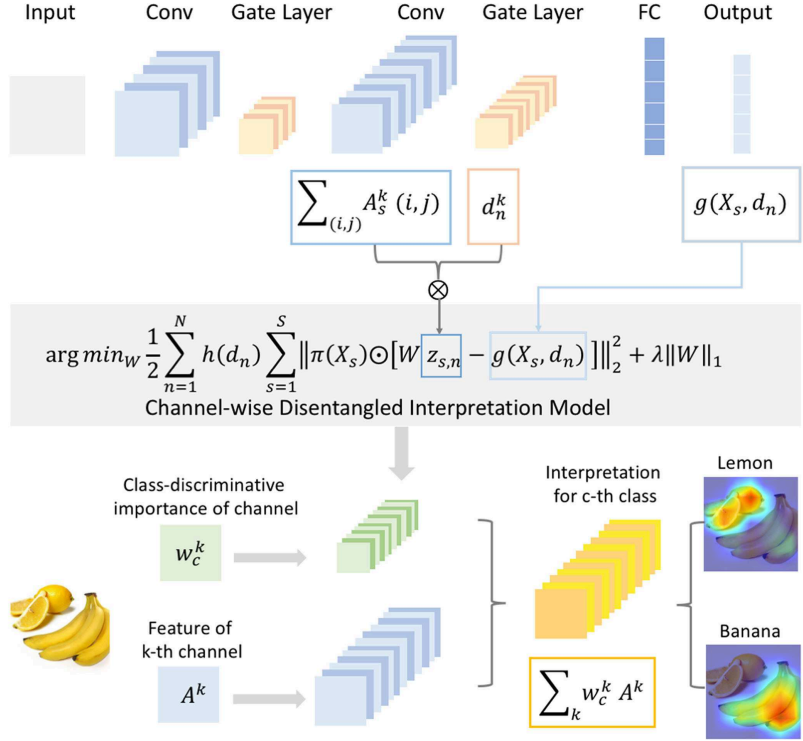}
	\centering
	\caption{The workflow of the proposed CHIP model. In the first stage, we learn the CHIP model based on the perturbed dataset which contains around a hundred million perturbed networks for the overall 1000 classes in ILSVRC 2015 dataset. After the optimization, we obtain the distilled class-discriminative importance of channels for different classes. In the second stage, given a specific image and the class of interest (banana and lemon), the class-discriminative visual interpretation is obtained for the target class by utilizing the distilled knowledge.}
	\label{fig:scheme1}
\end{figure*}
The perturbed network is generated by adding the channel gate layer after each original layer.
We denote the original features in the $l$-th layer as $\mathbf{A} \in \mathcal{R}^{u, v, K}$, where $u$ and $v$ are the width and height of a channel.
For the $k$-th channel in the $l$-th layer, the output of the channel gate layer is
\begin{equation}
\mathbf{\hat{A}}_{k} = d_k  \mathbf{A}_{k}
\end{equation}

For the $l$-th layer, the global average pooling of the channel gate layer is $\mathbf{z} = [z_1 ~ z_2 \cdots z_k \cdots z_{K}]^T$.
For the $k$-th channel, the global average pooling is 
\begin{equation}
z_k = \frac{1}{uv} \sum_{i,j} \mathbf{\hat{A}}_k(i,j)
\end{equation}

In the perturbed network, we add control gate layers behind each original layer without changing the original weight of the pre-trained network.

$\textbf{Perturbed dataset.}$ 
In order to learn the class-discriminative importance of channels, we need to generate the perturbed dataset, as shown in Fig. \ref{fig:illustration1}(bottom). 
Specifically, to learn $\mathbf{W}$ for the $l$-th layer, the perturbed dataset is obtained as follows: 
We first generate the perturbed networks. 
For the chosen $l$-th layer, we generate a channel gate sample $\mathbf{d}_n \in \{0,1\}^K$ by selecting elements of $\mathbf{d}_n$ as 1 uniformly at random.
Here, the number of the selected elements with value 1 is also uniformly sampled.
For other layers, we freeze the channel gates to be open.
We obtain the channel gate dataset $\mathcal{D}_{gate} = {\{ \mathbf{d}_n \}}_{n=1}^N$ by generating $N$ channel gate samples.

Secondly, we feed each image from image dataset $\mathcal{D}_{img} = {\{ \mathbf{X}_s \}}_{s=1}^S$ into each perturbed network and get the results.
For example, the result of the $n$-th perturbed network for the $s$-th image is denoted as 
\begin{equation}
g(\mathbf{X}_s,\mathbf{d}_n)=[g^1(\mathbf{X}_s,\mathbf{d}_n) \cdots g^c(\mathbf{X}_s,\mathbf{d}_n) \cdots g^C(\mathbf{X}_s,\mathbf{d}_n)]^{T} 
\end{equation}
where $g^c(\mathbf{X}_s,\mathbf{d}_n)$ is the prediction for the $c$-th class.
Meanwhile, its global average pooling in the $l$-th channel gate layer is recorded as $\mathbf{z}_{s,n}$.

Finally, we get the perturbed dataset $\mathcal{D} \lbrace\mathbf{X}_s, \mathbf{d}_n, \mathbf{z}_{s,n}, f(\mathbf{X}_s), g(\mathbf{X}_s, \mathbf{d}_n) \rbrace$, where the results of the original network are denoted as 
\begin{equation}
f(\mathbf{X}_s) = [f^1(\mathbf{X}_s) \cdots f^c(\mathbf{X}_s) \cdots f^C(\mathbf{X}_s)]^{T}
\end{equation}

\subsection{CHIP Model} \label{3b}
Given the perturbed dataset $\mathcal{D}$, we formulate the CHIP model as 
\begin{equation} \label{chip}
\begin{split}
\arg\min_{\mathbf{W}} & \frac{1}{2} \sum_{n=1}^{N} h(\mathbf{d}_{n}) \sum_{s=1}^{S} \|\pi(\mathbf{X}_s) \odot[\mathbf{W}\mathbf{z}_{s,n} - g(\mathbf{X}_s,\mathbf{d}_{n})]\|_2^{2} \\
& + \lambda \|\mathbf{W}\|_{1}
\end{split}
\raisetag{2\normalbaselineskip}
\end{equation}
where $\lambda$ is the regularization parameter, and $\odot$ denotes Hadamard product. The first term is the loss function, 
\begin{equation} \label{loss}
\mathcal{L} = \frac{1}{2} \sum_{n=1}^{N} h(\mathbf{d}_{n}) \sum_{s=1}^{S} \|\pi(\mathbf{X}_s) \odot[\mathbf{W}\mathbf{z}_{s,n} - g(\mathbf{X}_s,\mathbf{d}_{n})]\|_2^{2}
\end{equation}
 
In order to learn the class-discriminative importance weight, we approximate $\mathcal{L} $ by drawing perturbed samples which are weighted by $\pi(\mathbf{X}_s)$ and $h(\mathbf{d}_{n})$.

The perturbed dataset contains the data of different images from the image dataset $\mathcal{D}_{img} = {\{ \mathbf{X}_s \}}_{s=1}^S$.
To learn the class-discriminative importance, we denote $\pi(\mathbf{X}_s)$ as the loyalty measure of image $\mathbf{X}_s$ to different classes. 
For a given image, its loyalty for learning the class-discriminative importance of net-units to the $c$-th class is large when its $f^c(\mathbf{X}_s)$ is high.
Specifically, it is defined as
\begin{equation}
\pi(\mathbf{X}_s) = [\sqrt{f^1(\mathbf{X}_s)} \cdots \sqrt{f^c(\mathbf{X}_s)} \cdots \sqrt{f^C(\mathbf{X}_s)}]^{T}
\end{equation}
where $f^c(\mathbf{X}_s)$ denotes the network prediction that $\mathbf{X}_s$ belongs to the $c$-th class. 

The perturbed dataset also contains the data of different perturbed nets from the channel gate dataset $\mathcal{D}_{gate} = {\{ \mathbf{d}_n \}}_{n=1}^N$.
For a given perturbed net, $h(\mathbf{d})$ is denoted as the proximity measure between a binary channel gate vector $\mathbf{d}$ in the perturbed net and the all-one vector $\mathbf{1}$ in the original net. 
In the perturbed dataset, we sample perturbed nets both in the vicinity of the original net ( high $h(\mathbf{d}_{n})$ weight in Eq. (\ref{loss})) and at a distance from the original net (low $h(\mathbf{d}_{n})$ weight), where 
\begin{equation}
h(\mathbf{d}_{n}) = \exp(-\frac{1}{\sigma^{2}}\lVert \mathbf{d}_{n} - \mathbf{1} \rVert_{2}^{2}) .
\end{equation} 

The second term in our model is the sparse regularization term, 
\begin{equation}
\Phi(\mathbf{W}) = \|\mathbf{W}\|_{1}
\end{equation}
which measures the sparsity of the weight.
We use sparsity to measure the complexity of the interpretation model because of the inherent sparse property of features in DCNNs.
Also, the interpretation model needs to be simple enough to be interpretable.

Fig. \ref{fig:scheme1} illustrates the workflow of the proposed CHIP model. CHIP learns the class-discriminative importance of channels for different classes. The distilled knowledge can be further applied to visual interpretation and weakly-supervised object localization.

\begin{algorithm*} \label{algorithm}
\caption{Pseudocode of the CHIP Algorithm}
\LinesNumbered 

\KwIn{perturbed dataset $\mathcal{D} \lbrace\mathbf{X}_s, \mathbf{d}_n, \mathbf{z}_{s,n}, f(\mathbf{X}_s), g(\mathbf{X}_s, \mathbf{d}_n) \rbrace$; original net feature $\mathbf{A}$ for the instance being explained;}
\KwOut{optimal $\mathbf{W}^{\ast}$; visual interpretation obtained by CHIP for the given instance;} 
		
		 \textbf{Initialization:} set $c = 0$, $i = 0$, $\mathbf{m}_{c}^{0}$, $\mathbf{q}^{0}$, $\lambda > 0$, $\rho \geq 0$; \\
		  \Repeat () {class-discriminative importance of a specific layer for all categories are optimized}
		 {
		 The optimization problem for class-discriminative importance $\mathbf{w}_{c}$ of a specific layer for the $c$-th class;\\
		 $\arg\min_{\mathbf{w}_{c}} \frac{1}{2} \sum_{n=1}^{N} h(\mathbf{d}_{n}) \sum_{s=1}^{S} f^c(\mathbf{X}_s) [\mathbf{w}_{c}\mathbf{z}_{s,n} - g^c(\mathbf{X}_s,\mathbf{d}_{n})]^{2} + \lambda \|\mathbf{w}_{c}\|_{1}$; (Eq. (\ref{interpret1-2}))\\
		 \Repeat () {stopping criterion is satisfied}
		 {		 			
		1:~ $\mathbf{w}_{c}^{i+1} \leftarrow (\sum_{s,n} f^c(\mathbf{X}_s) h(\mathbf{d}_{n}) g^c(\mathbf{X}_s,\mathbf{d}_{n}) {\mathbf{z}_{s,n}}^{T} + \rho \mathbf{m}_{c}^i + \rho \mathbf{q}^i)(\sum_{s,n} f^c(\mathbf{X}_s) h(\mathbf{d}_{n}) \mathbf{z}_{s,n} {\mathbf{z}_{s,n}}^{T} + \rho \mathbf{I})^{-1}$; (Eq. (\ref{wc}))\\
		2:~ $\mathbf{m}_{c}^{i+1} \leftarrow soft (\mathbf{w}_{c}^{i+1} - \mathbf{q}^i, \frac{\lambda}{\rho})$; (Eq. (\ref{mc}))\\
		3:~ update lagrange multipliers:\quad $\mathbf{q}^{i+1} \leftarrow \mathbf{q}^{i} - (\mathbf{w}_{c}^{i+1} - \mathbf{m}_{c}^{i+1})$; (Eq. (\ref{q})) \\
		4:~ \textbf{update iteration:} \quad $i \leftarrow i+1$;\\		
		 }
		 \textbf{update iteration:} $c \leftarrow c+1$;\\	 
		 }
		 \textbf{return} $\mathbf{W}^{\ast}$;
		 CHIP interpretation of a specific layer for the target class: $\mathbf{\tilde{A}}^c = \sum_{k} {w_{c}^k} {\mathbf{A}_{k}}$; \\ 
\end{algorithm*} 

\textbf{CHIP Interpretation.}
By combining the class-discriminative importance and corresponding feature maps, we can get the class-discriminative visual interpretation for a specific network decision of an input image.
Specifically, for a certain layer, the visual interpretation for the $c$-th class prediction of a particular instance is denoted as
\begin{equation}
\mathbf{\tilde{A}}^c = \sum_{k} {w_{c}^k} {\mathbf{A}_{k}}
\end{equation}
where $w_c^k$ represents the optimal importance of the $k$-th channel to the $c$-th class, and ${\mathbf{A}_{k}}$ denotes the feature in the $k$-th channel for the given image.

\textbf{Refined CHIP Interpretation.}
The CHIP model can give the visual interpretation for a network decision.
Further, we design the Refined CHIP to present a high-resolution visual interpretation which can show detailed features distilled from the network.
It is commonly known that features in a shallow layer are of a higher resolution than that in a deep layer.
Conversely, the semantic representation in a deep layer is at a higher level than that in a shallow layer. 
Likewise, CHIP interpretation for shallow layers and deep layers also possess similar properties. 
We can combine the distilled interpretation in different layers to obtain the Refined CHIP visual interpretation.

Specifically, the Refined CHIP result is obtained by the point-wise multiplication of CHIP visual interpretations for the first and the last convolutional layers. 
Therefore, Refined CHIP interpretation is not only high-semantic but also high-resolution. 
It utilizes the distilled interpretation knowledge in different layers, which also reveals the roles of different layers. 
Recall that Guided Grad-CAM combines Guided Backpropagation visualization with Grad-CAM to give a high-resolution class-discriminative interpretation. 
For comparison, we also combine Guided Backpropagation \cite{springenberg2014striving} and CHIP interpretation to get Guided CHIP interpretation.

\textbf{Weakly-supervised Localization.}
It is well known that deep layers in DCNNs capture high-level semantic information which can be regarded as the object saliency information for localization. 
Therefore, the intuition is that our model distills the class-discriminative knowledge in the deep layer and transfers this knowledge from the pre-trained classification network to the localization task. 
Here, the visual interpretation can be regarded as a saliency map to localize the object. 
The bounding box can be obtained based on the saliency map distilled from the last convolutional layer. 
Because the visual interpretation obtained by our model is learned without the ground truth bounding box annotation, our localization method is also a weakly-supervised approach.

\subsection{CHIP Algorithm} \label{3c}
The proposed CHIP model is used to distill important channels of a pre-trained network for different image categories.
Here, we propose a channel-wise disentangled interpretation algorithm to optimize the CHIP model.

The optimization of CHIP model in Eq. (\ref{chip}) can be divided into solving the optimization problem for each class separately. 
For the $c$-th class, the optimization problem turns into
\begin{equation} \label{interpret1-2} 
\begin{split}
\arg\min_{\mathbf{w}_{c}} & \frac{1}{2} \sum_{n=1}^{N} h(\mathbf{d}_{n}) \sum_{s=1}^{S} f^c(\mathbf{X}_s) [\mathbf{w}_{c}\mathbf{z}_{s,n} - g^c(\mathbf{X}_s,\mathbf{d}_{n})]^{2} \\
& + \lambda \|\mathbf{w}_{c}\|_{1}
\end{split}
\raisetag{2\normalbaselineskip}
\end{equation}

The optimization problem in Eq. (\ref{interpret1-2}) is convex.
Here, we design a channel-wise disentangled interpretation algorithm by adopting the alternating iteration rule to learn $\mathbf{w}_{c}$.

The optimization problem can be converted into the equivalent formulation
\begin{equation}
\begin{split}
\arg\min_{\mathbf{w}_{c},\mathbf{m}_{c}}& \frac{1}{2} \sum_{n=1}^{N} h(\mathbf{d}_{n}) \sum_{s=1}^{S} f^c(\mathbf{X}_s) [\mathbf{w}_{c}\mathbf{z}_{s,n} - g^c(\mathbf{X}_s,\mathbf{d}_{n})]^{2} \\
&+ \lambda \|\mathbf{m}_{c}\|_{1} \\
\text{subject to}& \quad \mathbf{w}_{c} = \mathbf{m}_{c}
\end{split}
\raisetag{2\normalbaselineskip}
\end{equation}

The augmented Lagrangian for the above problem is
\begin{equation}
\begin{split}
\arg\min_{\mathbf{w}_{c}, \mathbf{m}_{c}, \mathbf{p}} &\frac{1}{2} \sum_{n=1}^{N} h(\mathbf{d}_{n}) \sum_{s=1}^{S} f^c(\mathbf{X}_s) [\mathbf{w}_{c}\mathbf{z}_{s,n} - g^c(\mathbf{X}_s,\mathbf{d}_{n})]^{2} \\
&+ \lambda \|\mathbf{m}_{c}\|_{1} + \mathbf{p}^T (\mathbf{w}_{c} - \mathbf{m}_{c}) + \frac{\rho}{2} \Vert \mathbf{w}_{c} - \mathbf{m}_{c} \Vert_2^2
\end{split}
\raisetag{2\normalbaselineskip}
\end{equation}

The equation can be rewritten as 
\begin{equation}
\begin{split}
\arg\min_{\mathbf{w}_{c}, \mathbf{m}_{c}, \mathbf{q}} &\frac{1}{2} \sum_{n=1}^{N} h(\mathbf{d}_{n}) \sum_{s=1}^{S} f^c(\mathbf{X}_s) [\mathbf{w}_{c}\mathbf{z}_{s,n} - g^c(\mathbf{X}_s,\mathbf{d}_{n})]^{2}\\
&+ \lambda \|\mathbf{m}_{c}\|_{1} + \frac{\rho}{2} \Vert \mathbf{w}_{c} - \mathbf{m}_{c} - \mathbf{q} \Vert_2^2
\end{split}
\raisetag{2\normalbaselineskip}
\end{equation}
where 
\begin{equation}
\mathbf{q} \equiv - \frac{1}{\rho} \mathbf{p}
\end{equation}

Through a careful choice of the new variable, the initial problem is converted into a simple problem.
Given that the optimization is considered over the variable $\mathbf{w}_{c}$, the optimization function can be reduced to
\begin{equation}
\begin{split}
\mathbf{w}_{c} \leftarrow \arg\min_{\mathbf{w}_{c}} &\frac{1}{2} \sum_{n=1}^{N} h(\mathbf{d}_{n}) \sum_{s=1}^{S} f^c(\mathbf{X}_s) [\mathbf{w}_{c}\mathbf{z}_{s,n} - g^c(\mathbf{X}_s,\mathbf{d}_{n})]^{2}\\
&+ \frac{\rho}{2} \Vert \mathbf{w}_{c} - \mathbf{m}_{c} - \mathbf{q} \Vert_2^2
\end{split}
\raisetag{2\normalbaselineskip}
\end{equation}

The solution is
\begin{equation} \label{wc}
\begin{split}
\mathbf{w}_{c}^{i+1} \leftarrow &(\sum_{s,n} f^c(\mathbf{X}_s) h(\mathbf{d}_{n}) g^c(\mathbf{X}_s,\mathbf{d}_{n}) {\mathbf{z}_{s,n}}^{T} + \rho \mathbf{m}_{c}^i \\
&+ \rho \mathbf{q}^i)(\sum_{s,n} f^c(\mathbf{X}_s) h(\mathbf{d}_{n}) \mathbf{z}_{s,n} {\mathbf{z}_{s,n}}^{T} + \rho \mathbf{I})^{-1}
\end{split}
\raisetag{2\normalbaselineskip}
\end{equation}

To calculate $\mathbf{m}_{c}$, the optimization problem to be solved is
\begin{equation}
\mathbf{m}_{c} \leftarrow \arg \min_{\mathbf{m}_{c}} \lambda \Vert \mathbf{m}_{c} \Vert_1 + \frac{\rho}{2} \Vert \mathbf{w}_{c} - \mathbf{m}_{c} - \mathbf{q} \Vert_2^2
\end{equation}

The solution is
\begin{equation} \label{mc}
\mathbf{m}_{c}^{i+1} \leftarrow soft (\mathbf{w}_{c}^{i+1} - \mathbf{q}^i, \frac{\lambda}{\rho})
\end{equation}

Lagrange multipliers update to
\begin{equation} \label{q}
\mathbf{q}^{i+1} \leftarrow \mathbf{q}^{i} - (\mathbf{w}_{c}^{i+1} - \mathbf{m}_{c}^{i+1})
\end{equation}

The pseudocode of the CHIP algorithm is shown in Algorithm \ref{algorithm}.

\section{Experiments}
\label{sec:Exp}  
In this section, we evaluate the performances of the CHIP model and the Refined CHIP interpretation.
We first describe experiment settings of our interpretation model in section \ref{setting}.
We then conduct experiments in five aspects to qualitatively and quantitatively evaluate the proposed model.

Specifically, in section \ref{visual}, we show the CHIP and the Refined CHIP visual interpretation of the VGG16 network \cite{simonyan2014very} for object classification based on the ILSVRC 2015 object classification dataset \cite{Russakovsky2015}.
In section \ref{loc}, the performance of the CHIP model on the VGG16 network for object classification is quantitatively evaluated based on the ILSVRC 2015 object localization dataset.
In section \ref{visual-scene}, we qualitatively and quantitatively evaluate the performance of the proposed model on the Inception-V3 network \cite{szegedy2016rethinking} for scene classification task based on the ADE20K dataset \cite{8100027}. 
In section \ref{layers}, we show the visual interpretation for different layers of the network. 
In section \ref{classes}, the class-discriminative importance of channels for different classes are compared.

\subsection{Experiment Settings} \label{setting}
 
\subsubsection{Networks Being Interpreted}
The CHIP model is applied to interpret the decisions of VGG16 and Inception-V3.

\textbf{VGG16 on the ILSVRC 2015 object classification dataset:}
To be consistent with the previous work, we use the off-the-shelf VGG16 model from the Caffe Model Zoo as one network to be interpreted.
VGG16 involves 16 layers with learnable weight. 
Different layers have different number of channels.
For example, the first convolutional layer contains 64 channels while the last convolutional layer has 512 channels.
The object classification dataset contains 1000 object classes.

\textbf{Inception-V3 on the Places365 scene classification dataset:} Inception-V3 is a DCNN that belongs to the GoogLeNet series. 
The network is 48 layers deep and can learn rich feature representations. 
Places365 scene classification dataset involves 365 scene classes \cite{zhou2017places}.

\subsubsection{Quantitative Evaluation} The following human-annotated datasets are used for quantitative evaluation.

\textbf{ILSVRC 2015 object localization dataset:}
To quantitatively evaluate the CHIP model for VGG16 on object classification, we utilize the ILSVRC 2015 object localization dataset and the Intersection over Union (IoU) metric.
In this dataset, images are labeled with ground-truth classes and corresponding bounding boxes.

\textbf{ADE20K scene understanding dataset:}
To quantitatively evaluate the CHIP model for Inception-V3 on scene classification, we utilize the ADE20K scene understanding dataset with pixel-wise annotations.
ADE20K and Places365 have many overlapped scene classes.
Therefore, these overlapped scene classes in ADE20K with pixel-wise annotations can be used to assess the CHIP interpretation for Inception-V3 on Places365.
ADE20K contains object segmentation annotation for each image.
Based on the IoU metric, quantitative evaluation is computed by comparing the CHIP interpretation with the critical object annotation for the target scene class.

In the quantitative comparison, the IoU metric is defined as:
\begin{equation}
\textbf{IoU} = \dfrac{\textit{Area of Overlap}}{\textit{Area of Union}} 
\end{equation}
where the numerator refers to the area of overlap between the CHIP interpretation region and the ground-truth region for the target category, and the denominator denotes the area of the union of the two regions. 

\subsubsection{The Learning of CHIP Model}

\begin{figure*}[htp]
	\centering
	\includegraphics[width=0.9\linewidth]{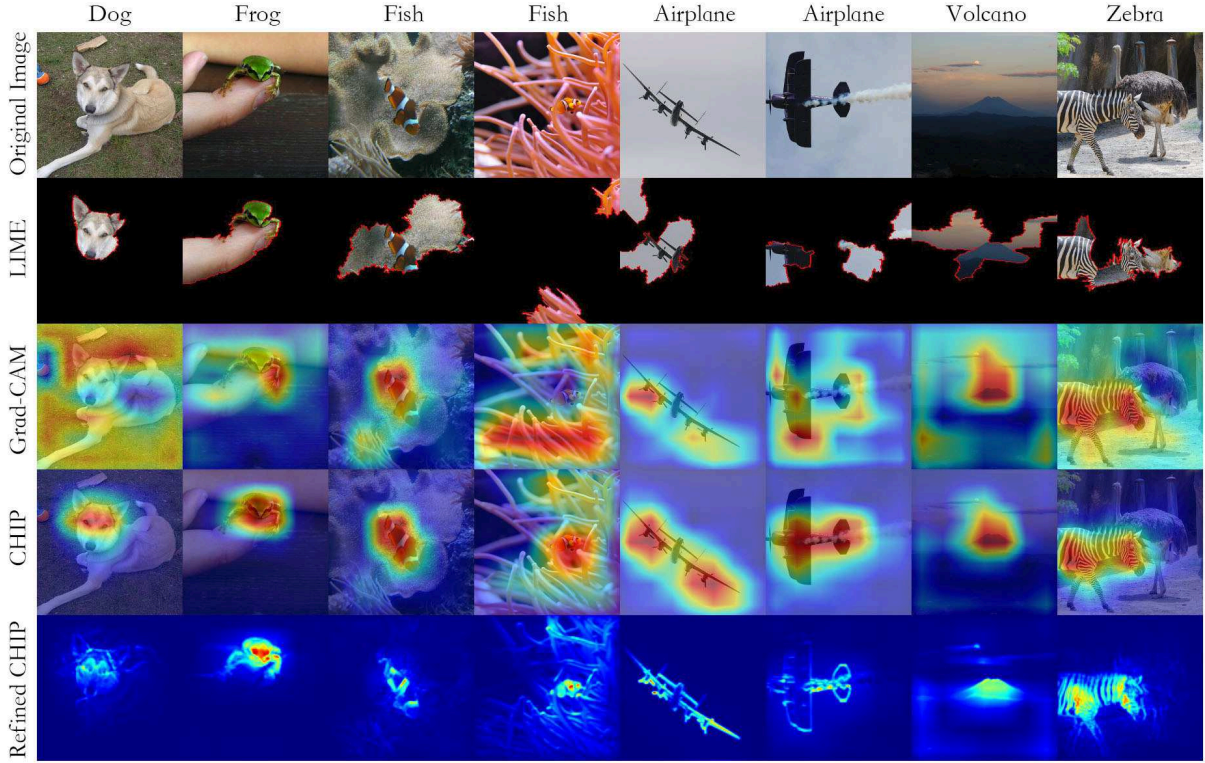}
	\centering
	\caption{Comparison of visual interpretations for simple one-object images.}
	\label{fig:feature_map_simple}
\end{figure*}

By adding gate layers on top of each original layer in the network being interpreted, we can build perturbed networks for the learning of our interpretation model.

In order to learn the class-discriminative importance of channels in a specific layer for different classes, we generate $100 \times Image\, Number\, in\, Dataset$ perturbed networks.
For the interpretation of VGG16 on the ILSVRC 2015 object classification dataset, we learn the CHIP model by using the training dataset which includes about 1.3 million images of total 1000 classes.
For each class, the number of images ranges from 732 to 1300.
Therefore, for the 1000 object classes, the number of perturbed networks is around a hundred million. 
In each perturbed network, we only perturb the layer of interest, while keeping other gate layers in open states.
We get the perturbed dataset by feeding images into perturbed networks, as described in Section \ref{3a}.
After the optimization, we can get the class-discriminative importance of channels in the layer of interest for each class.

\subsection{Visual Interpretation for object classification} \label{exp1}
\begin{figure*}[htp]
	\centering
	\includegraphics[width=0.7\linewidth]{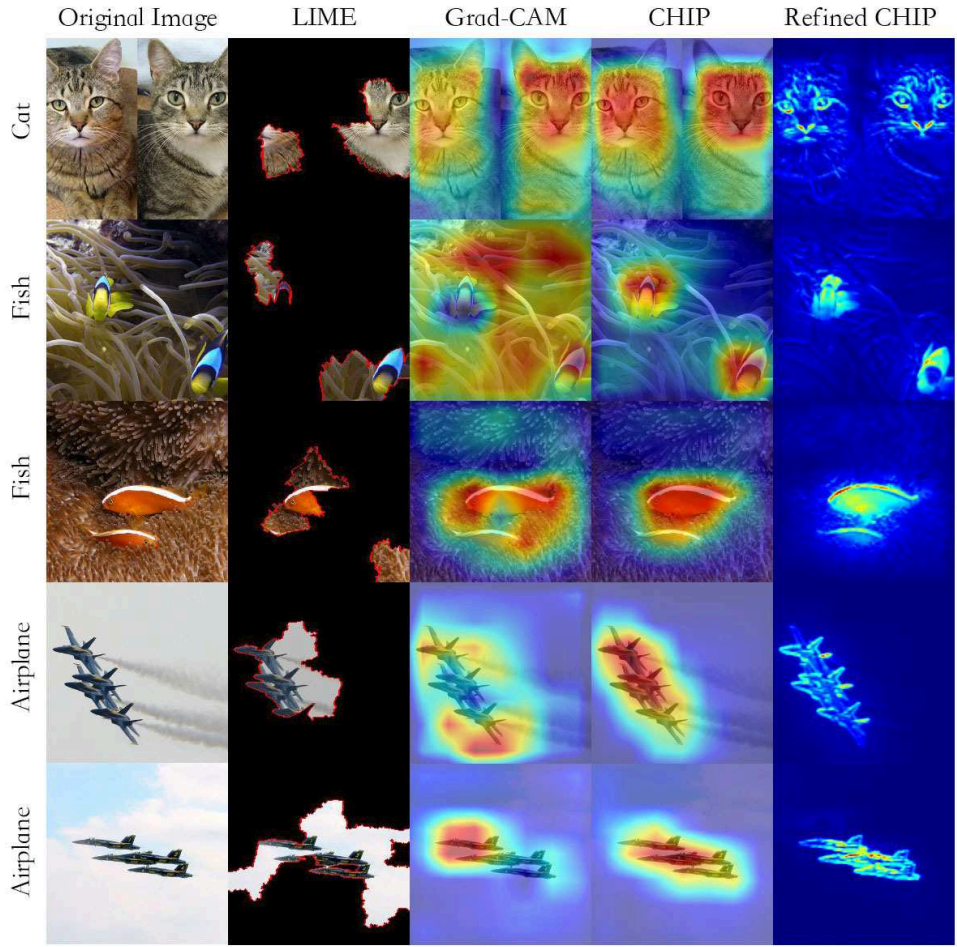}
	\centering
	\caption{Comparison of visual interpretations for complex multi-object images.}
	\label{fig:feature_map_complex}
\end{figure*}  
In this section, we qualitatively and quantitatively evaluate the visual interpretation performances of CHIP and Refined CHIP for VGG16 in object classification task. We learn class-discriminative importance of channels in the first and the last convolutional layers for 1000 classes.
Qualitatively, the CHIP model provides visual interpretation results to explain network predictions.
Quantitatively, it is evaluated by IoU in weakly-supervised localization task. 

\subsubsection{\textbf{Qualitative evaluation on the ILSVRC 2015 object classification dataset}}\label{visual}
For qualitative evaluation, we choose images from ILSVRC 2015 object classification validation dataset as the test images.

Previous work has demonstrated that the semantic representation in a deep convolutional layer is more class-specific than that in a shallow convolutional layer. 
In light of that, CHIP model focuses on the class-discriminative visual interpretation in the last convolutional layer.
Meanwhile, features in a shallow layer are of a higher resolution than that in a deep layer.
Therefore, the distilled interpretation in the first and the last convolutional layers can be combined to obtain the Refined CHIP interpretation which is both high-semantic and high-resolution. 
We compare the CHIP and Refined CHIP with LIME \cite{Ribeiro2016}, Grad-CAM \cite{8237336}, Guided Backpropagation \cite{springenberg2014striving} and Guided Grad-CAM \cite{8237336}.
LIME and Grad-CAM can only present coarse class-discriminative visual interpretation.

Fig. \ref{fig:feature_map_simple} shows the comparison of visual interpretations of different methods for simple images, each of which only contains a single target of the selected class. E.g., the first original image contains a dog, and visual interpretation results of different methods are provided for the dog class.
For LIME, the class-discriminative importance is calculated for each super-pixel.
Then, LIME explains the prediction by selecting image regions that is important to the target class.
In the experiment, LIME shows the top 5 important super-pixels for the target class.
For Grad-CAM and CHIP, the visual interpretation is designed at the pixel level, which can also be regarded as the saliency map for the target class.
Similar to the previous work, visual interpretations of Grad-CAM and CHIP are displayed by superimposing saliency maps on the original image.
Because the results of Grad-CAM and CHIP are not high-resolution, this displaying way makes it easier to evaluate whether the important pixels in visual interpretation belong to the object region of the target class. 
However, this displaying way is not needed for Refined CHIP because it is high-resolution interpretation containing fine-scale features. Therefore, we directly show the interpretation results of Refined CHIP.

From Fig. \ref{fig:feature_map_simple}, we can observe that LIME, in most cases, is able to capture partial target object as the visual interpretation, except for the fourth original image where a fish is hiding near coral.
The flaw of LIME is that its visual interpretation is limited by the image segmentation result.
LIME also cannot explore inside the network, which restricts its performance.
Its results usually contain image region that does not belong to the target object, such as the sixth LIME explanation where the sky is included as the critical image region for the airplane.

Compared with LIME and Grad-CAM, CHIP is more reasonable since it always captures the target object.
For instance, in the explanation of the fourth original image, LIME and Grad-CAM both miss the fish object region and highlight the background, while the proposed CHIP and Refined CHIP both highlight the correct object region.
It illustrates that CHIP and Refined CHIP have better class-discriminative characteristics.
On top of CHIP, Refined CHIP presents fine-grained details for the target object, which further illustrates the inherent characteristics of features in different layers.
For example, for the last original image in Fig. \ref{fig:feature_map_simple}, Refined CHIP depicts the zebra-stripes in the visual interpretation for zebra.
In contrast, other compared methods can only provide coarse visual interpretation, such as the super-pixel in LIME and the salient region in Grad-CAM. 

Fig. \ref{fig:feature_map_complex} shows the visual interpretation in complex images each of which contains multiple objects of the target class.
In this situation, a good visual interpretation is expected to highlight all objects of target category, which is more difficult.
In this figure, the compared methods neglect some target objects in some cases.
As shown in the second row of Fig. \ref{fig:feature_map_complex}, LIME and Grad-CAM both miss one fish.
In contrast, CHIP provides more comprehensive visual interpretation which captures all objects of the target category.
Refined CHIP further shows more detailed visual interpretation for the target category, such as the explanation for the fourth original image where the contour and texture of airplanes are presented.

\begin{figure*}[htp]
	\centering
	\includegraphics[width=0.9\linewidth]{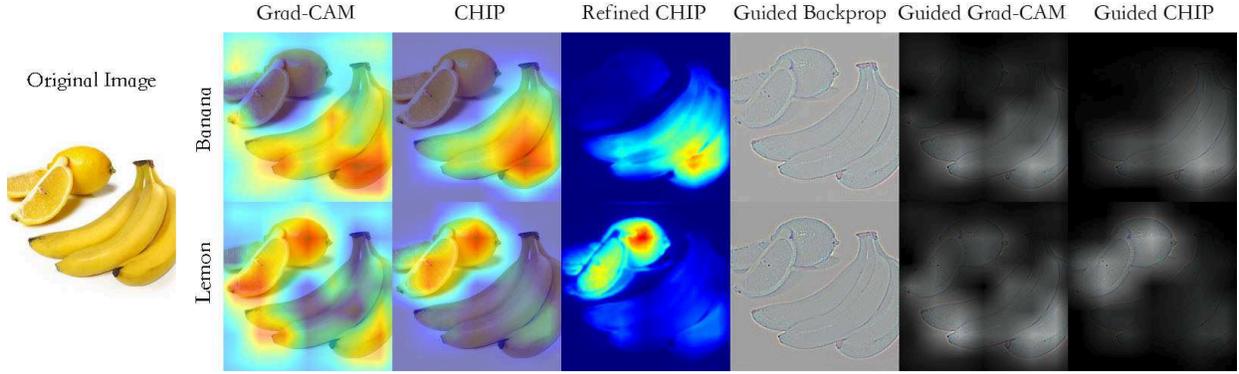}
	\centering
	\caption{Comparison of visual interpretations for a multi-category image.}
	\label{fig:feature_map_GB}
\end{figure*}

Fig. \ref{fig:feature_map_GB} shows the visual interpretation for an image where there are multiple target categories.  
The target categories are lemon and banana in the original image.
The visual interpretation for banana and lemon are respectively in the top and bottom half of Fig. \ref{fig:feature_map_GB}.
In this figure, we also compare Refined CHIP with Guided Backpropagation, Guided Grad-CAM, and Guided CHIP.

Fig. \ref{fig:feature_map_GB} indicates that CHIP performs more reasonable than Grad-CAM.   
Grad-CAM does not provide a reasonable explanation for lemon, because it also highlights partial region of banana.
Fig. \ref{fig:feature_map_GB} shows Guided Backpropagation provides high-resolution but not class-discriminative visualization, because its visualization results for lemon and banana are similar.
The visual interpretation of Guided Grad-CAM and Guided CHIP involve rich details by combining Guided Backpropagation.
However, because of the inferior visual interpretation of Grad-CAM, Guided Grad-CAM is also less reasonable than Guided CHIP.
As shown in Fig. \ref{fig:feature_map_GB}, Guided Grad-CAM for lemon highlights banana region, while Guided CHIP only highlights lemon region. 
In Fig. \ref{fig:feature_map_GB}, Refined CHIP presents visual interpretation that is both high-resolution and class-discriminative without using Guided Backpropagation.
Except for the correctly highlighted object region, the fine-grained information in Refined CHIP further identifies the important fine-scale features in the network for the predicted class.

\begin{table*}
  \caption{Localization and classification errors on ILSVRC 2015 validation set.}
  \label{cmp_table}
  \centering
  \begin{tabular}{c||c|c||c|c|c}
    \hline\hline
    Localization method &Top 1 loc error &Top 5 loc error &Classification network &Top 1 cls error &Top 5 cls error \\

    \hline

    Backpropagation \cite{simonyan2013deep} &61.11 &51.43 & & &  \\
     & & & & &  \\
    c-MWP \cite{Zhang2018} &70.92 &63.01 &  & &  \\
     & & &VGG16 &30.12 &10.85  \\
    Grad-CAM \cite{8237336} &56.47 &46.35 &  & &  \\
     & & & & &  \\
    CHIP &\textbf{51.45} &\textbf{40.16} & & & \\

    \hline\hline
  \end{tabular}
\end{table*}
 
\subsubsection{\textbf{Quantitative evaluation on the ILSVRC 2015 object localization dataset}} \label{loc}
In this experiment, we assess the CHIP model by the ILSVRC 2015 object localization dataset, the ground-truth of which is human-annotated bounding-boxes.
In total, there are 50000 images for the 1000 classes in the validation dataset. 
The network being interpreted is the VGG16 on the ILSVRC 2015 object classification dataset. 

In weakly-supervised object localization task, competing methods should localize the target object without the ground-truth localization annotation.
In this experiment, competing approaches are evaluated in the ILSVRC 2015 weakly-supervised localization task where they are required to provide object bounding boxes together with classification predictions. 
Specifically, given an input image, the original classification network gives its class prediction.
And different competing methods are used to learn the saliency map from the interpretation of the classification network for the predicted class. 
The obtained saliency map is binarized with the optimal threshold of the maximal intensity.
The corresponding bounding box is obtained around the largest partition in the binarized saliency map.
For each competing method, a grid search is implemented to select the optimal threshold for the best localization performance.
Finally, the compared methods can get the bounding boxes as the localization results.
Here, quantitative and qualitative analyses are provided to evaluate the competing methods.

\begin{figure*}[htp]
	\centering
	\includegraphics[width=1\linewidth]{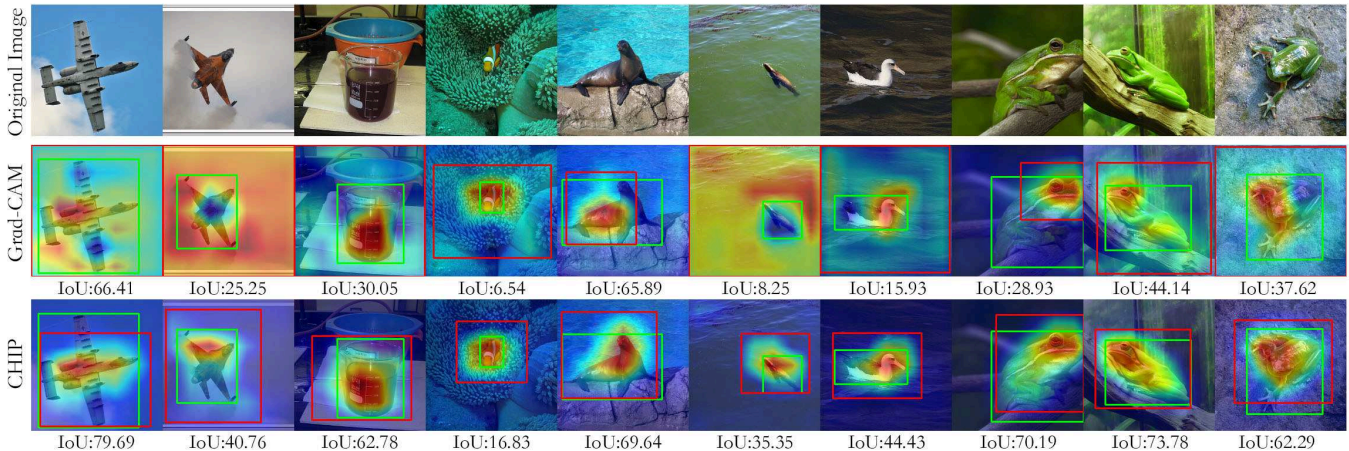}
	\centering
	\caption{Visual examples depicting the object localization performance of Grad-CAM and CHIP. The green boxes are ground-truth annotations for images and red boxes are predicted bounding boxes. IoU is shown under each interpretation result.}
	\label{fig:loc_map}
\end{figure*} 

For the quantitative comparison, we evaluate the localization error using the IoU metric in which the numerator refers to the area of overlap between the predicted bounding box and the ground-truth bounding box for the target category, and the denominator denotes the area of their union. 

Table \ref{cmp_table} shows the localization errors of competing methods in the ILSVRC 2015 validation dataset. 
Here, the proposed method is compared with Grad-CAM \cite{8237336}, c-MWP\cite{Zhang2018}, and Backpropagation methods \cite{simonyan2013deep}.
To evaluate the pre-trained VGG16 neural network in the context of image classification, both top 1 and top 5 classification errors in validation dataset are reported in Table \ref{cmp_table}.
Because all competing methods are based on the same pre-trained VGG16 network, their classification accuracies are the same.
Table \ref{cmp_table} shows both top 1 and top 5 localization errors in validation dataset.
Table \ref{cmp_table} indicates that the localization error of our model is lower than other methods.
It also indicates that our model provides better class-discriminative saliency maps than other visual interpretation models in this task.

Fig. \ref{fig:loc_map} shows the qualitative and quantitative comparisons of Grad-CAM and the proposed approach.
Because Grad-CAM achieves better object localization result than c-MWP and Backpropagation, here we only show the result of Grad-CAM for comparison.
It can be seen that the object localization of CHIP with high IoU outperforms that of Grad-CAM.
As shown in Fig. \ref{fig:loc_map}, the predicted bounding boxes of Grad-CAM often cover the whole image region, such as the left three images.

The intuitive reason is that the saliency map of Grad-CAM sometimes highlights the background region rather than the target object, and thus results in the inaccurate localization.
In comparison, the predicted bounding boxes of the proposed CHIP method have a better overlap with ground-truth annotations, resulting in a lower localization error.
 
\subsection{Visual Interpretation for scene classification} \label{visual-scene}
In this section, we assess the performance of the CHIP model on the ADE20K scene understanding dataset.
The ground-truth of each image in ADE20K is a fine-scale human-annotated label.
Therefore, the experiment on this dataset provides an evaluation of the CHIP model by testing whether the visual interpretation of CHIP model matches human knowledge. 
From this point of view, the CHIP model is evaluated by the subjective assessment in this experiment.

Specifically, the CHIP model is applied to interpret the outputs of Inception-V3 network on the Places365 scene classification dataset.
To quantitatively evaluate its performance, we adopt the ADE20K scene understanding dataset in which the critical object for the scene class is labeled by the pixel-wise annotation.
In the quantitative evaluation, the IoU metric is measured by comparing the CHIP interpretation with the critical object annotation for the target scene class.

For each testing image, Inception-V3 network outputs its scene classification prediction.
Different competing interpretation methods are applied to explain the network decision.
The proposed CHIP model can distill the saliency map of the critical object for the target scene class. 
The interpretation results of the compared methods are obtained by binarizing saliency maps with the corresponding optimal threshold of the maximal intensity.

\begin{figure*}[htp]
	\centering
	\includegraphics[width=0.9\linewidth]{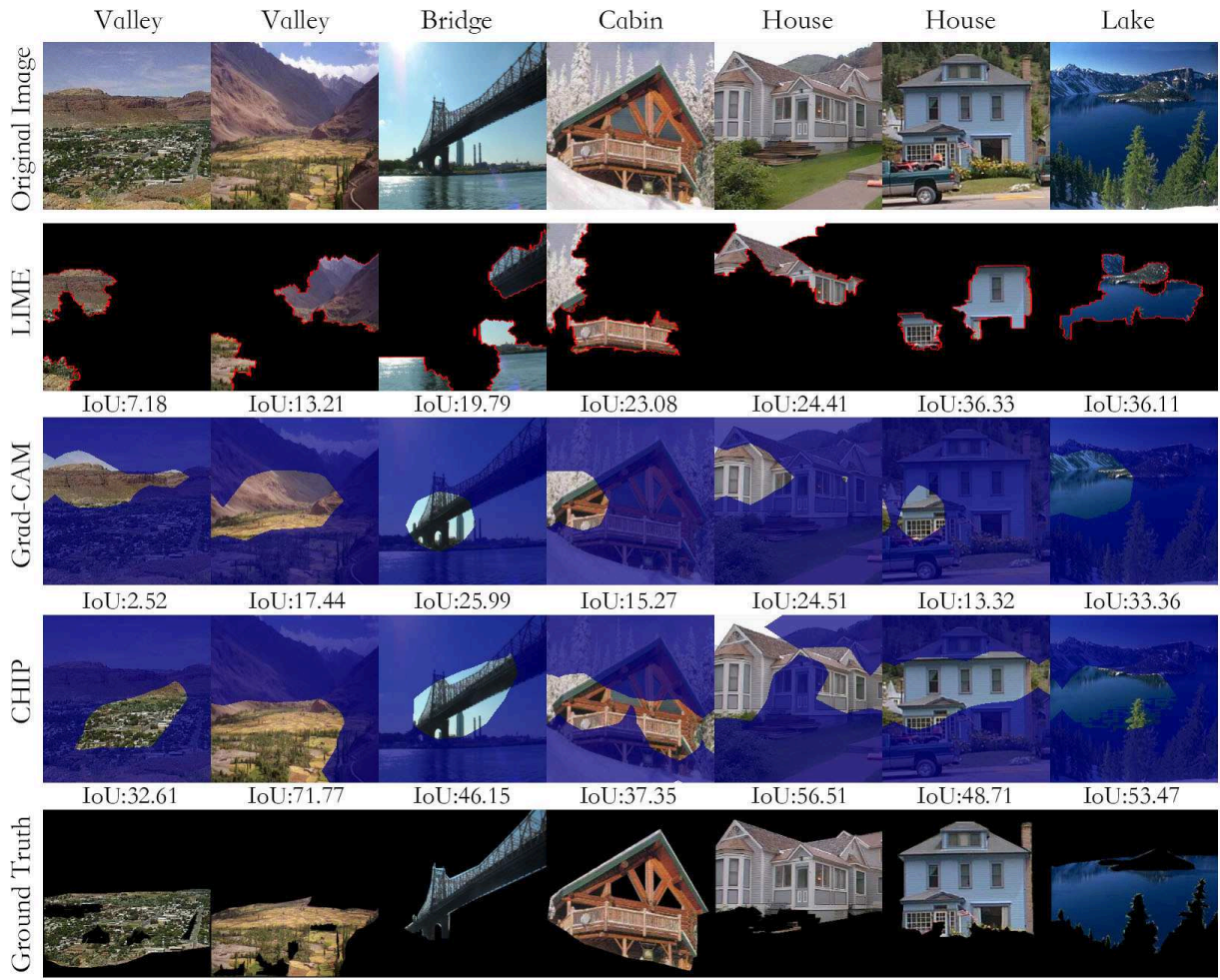}
	\centering
	\caption{Examples depicting visual interpretations of LIME, Grad-CAM, and CHIP for images from ADE20K. The highlighted patches in each row denote critical regions obtained by the compared methods to interpret the Inception-V3 prediction. IoU is shown under each interpretation result.}
	\label{fig:imgs_ade}
\end{figure*}

Fig. \ref{fig:imgs_ade} shows the qualitative and quantitative evaluations of CHIP, when compared with LIME, Grad-CAM and the ground truth of the critical object for the target scene class. We can see that CHIP outperforms the compared methods.
For example, in terms of the interpretation of valley in Fig. \ref{fig:imgs_ade}, Grad-CAM and LIME select the mountain region as the interpretation with low IoUs. 
In comparison, CHIP can interpret the net prediction with higher IoUs by distilling the valley class-discriminative knowledge from the network.

\subsection{Visual Interpretation at Different Layers} \label{layers}
\begin{figure*}[htp]
	\centering
	\includegraphics[width=0.6\linewidth]{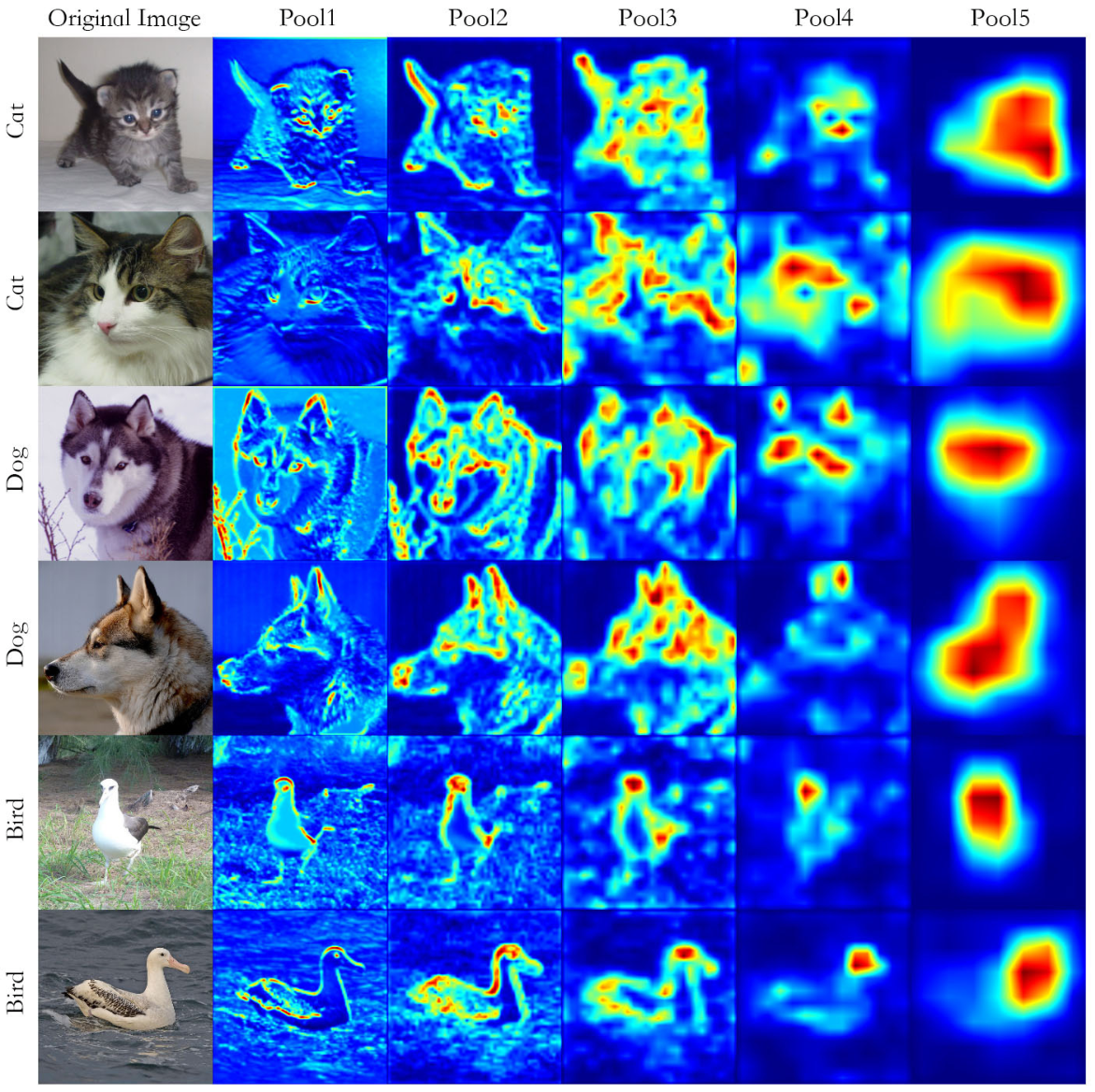}
	\centering
	\caption{Visual interpretations in different layers of VGG16 model for simple images.}
	\label{fig:5layer_simple}
\end{figure*}

\begin{figure*}[htp]
	\centering
	\includegraphics[width=0.6\linewidth]{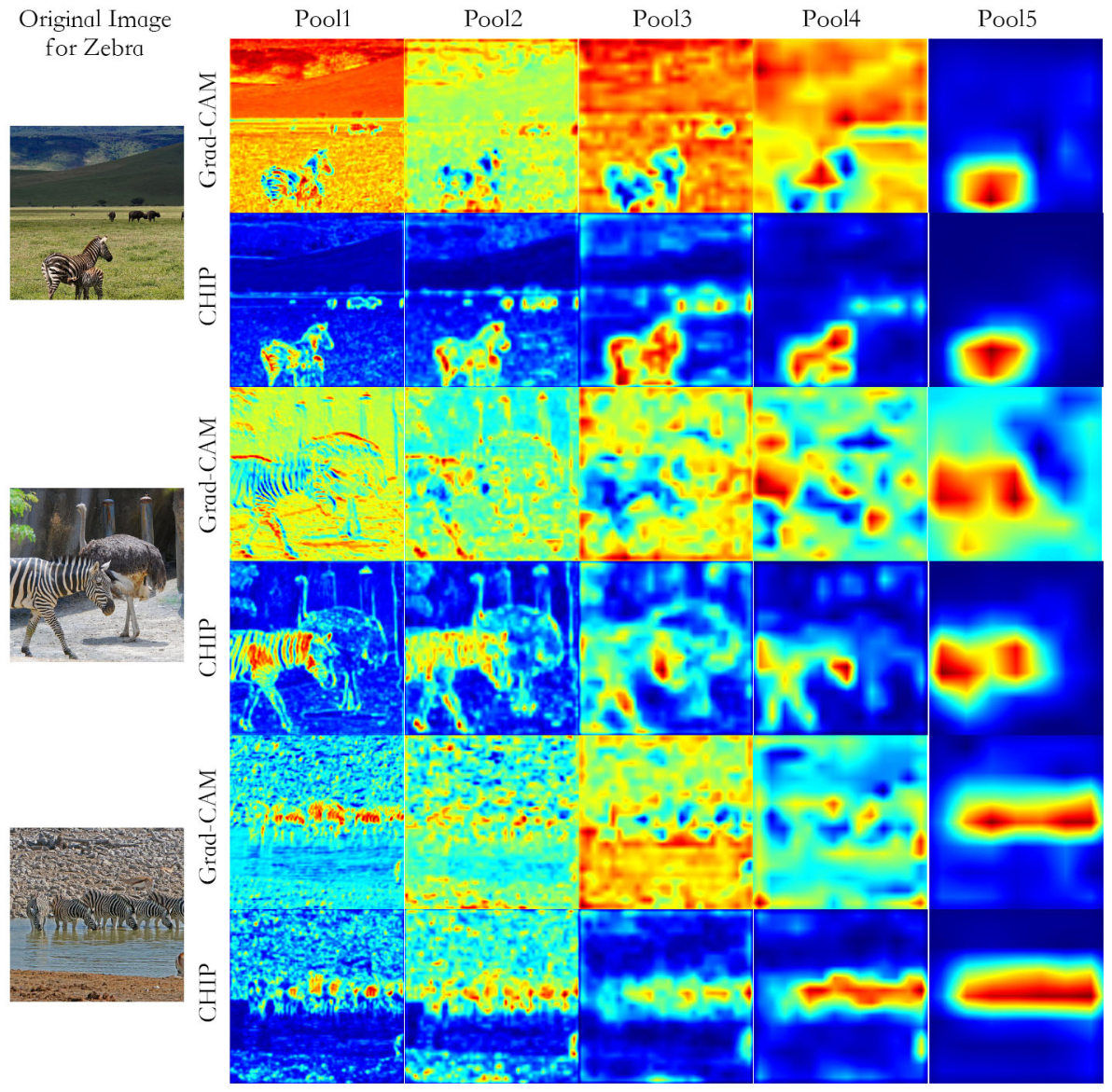}
	\centering
	\caption{Visual interpretations in different layers of VGG16 model for complex images.}
	\label{fig:5layers_complex}
\end{figure*}

In this section, we conduct experiments to show visual interpretations in different layers.
We select five pooling layers from VGG16 as the target layers.
For each pooling layer, the class-discriminative importance of channels for different classes are learned by the CHIP model.
Visual interpretations are then obtained by combining the learned importance of channels with the corresponding features. We choose images from ILSVRC 2015 validation dataset as test images.

Fig. \ref{fig:5layer_simple} shows visual interpretations for simple images. 
Here, "simple" means the object of target class is single and can be easily distinguished from background.
In Fig. \ref{fig:5layer_simple}, we select images from three classes, cat, dog, and bird.
Visual interpretations in the pool1 layer highlight fine-grained object details of the target classes, such as edges and textures shown in the interpretation for cat.
In contrast, interpretations in deeper layers capture higher-level semantic features of objects, such as eyes and ears of dog shown in the interpretation of the pool4 layer.
This observation is consistent with the previous visualization work \cite{6910029}.
We note that interpretations in different layers highlight the image region that is specific to the class of interest, which qualitatively demonstrates the effectiveness of the class-discriminative property of CHIP.

\begin{figure}[htp]
	\centering
	\includegraphics[width=0.9\linewidth]{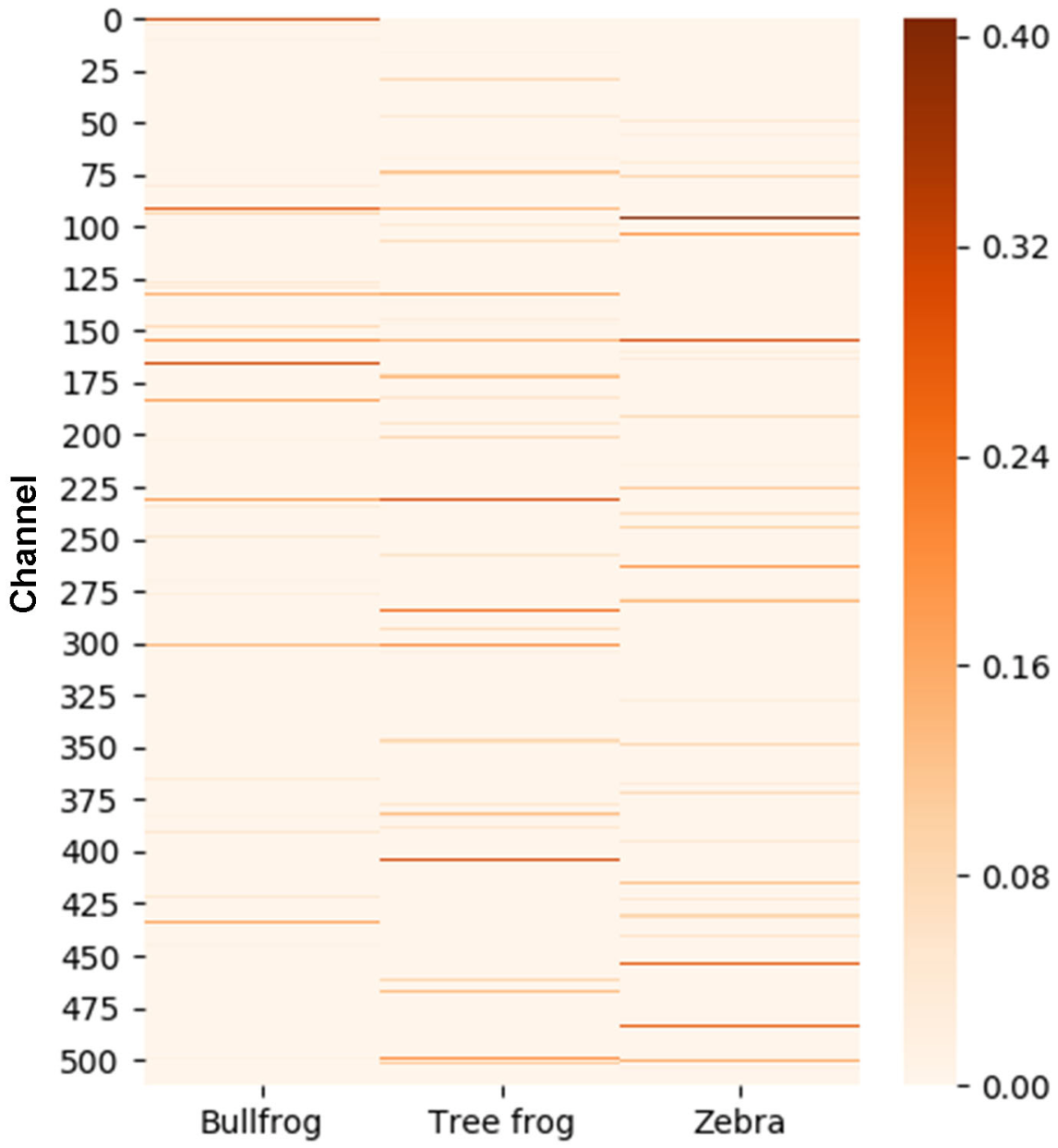}
	\centering
	\caption{Comparison of the class-discriminative importance of channels for different classes.}
	\label{fig:sparse}
\end{figure}

\begin{figure}[htp]
	\centering
	\includegraphics[width=1\linewidth]{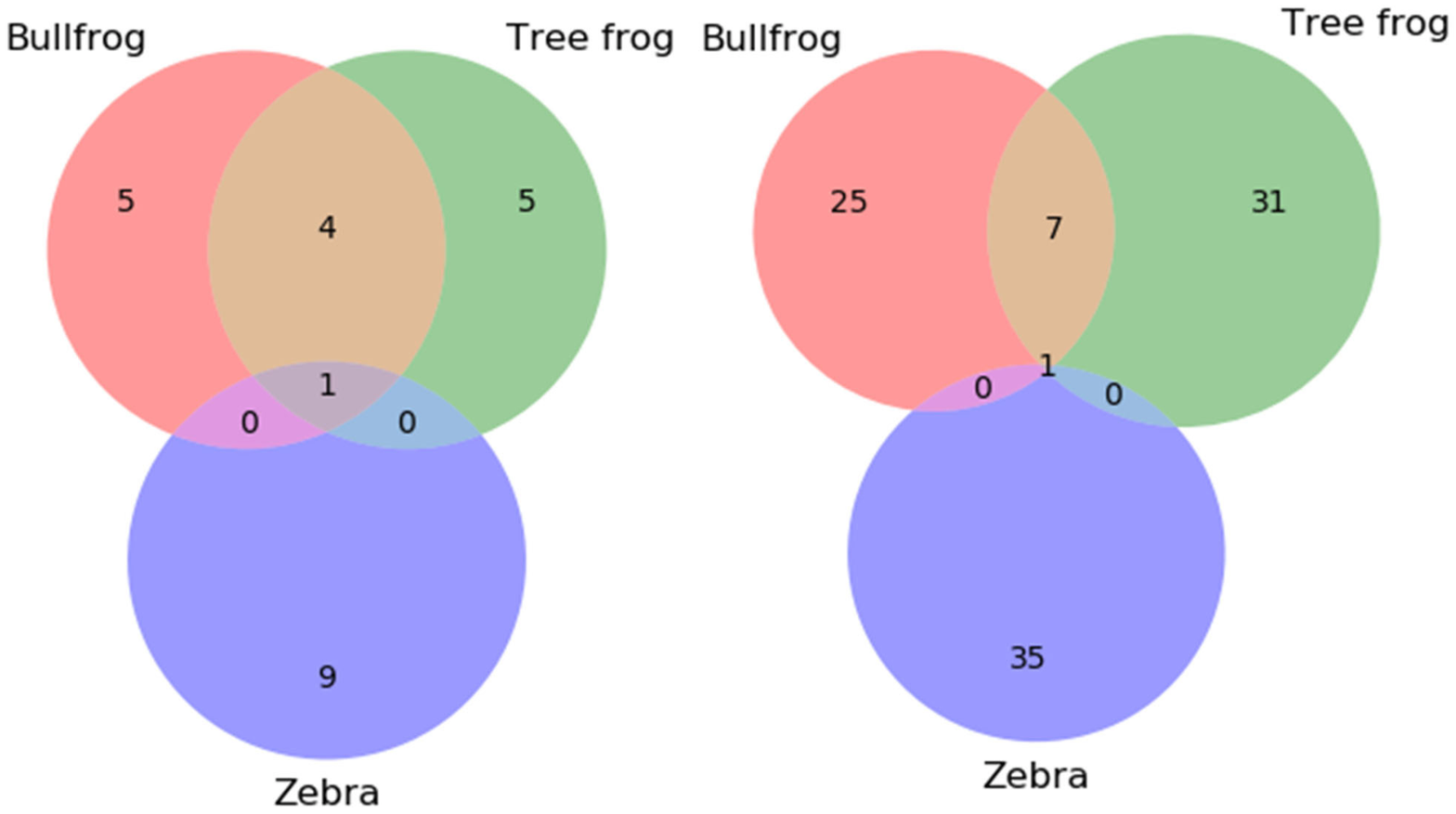}
	\centering
	\caption{Venn diagrams of the overlapped channels between different classes. The left shows the number of overlapped channels within the top 10 important channels. The right shows the number of overlapped channels whose importance are larger than a thousandth of the highest one for each class.}
	\label{fig:venn3}
\end{figure}

Fig. \ref{fig:5layers_complex} compares our visual interpretation with Grad-CAM in different layers for zebra class in complex images.
Here, "complex" means there are multiple objects of different classes in a complex background. E.g., the first original image has two zebras and some bisons in the grassland. 

In Grad-CAM, visual interpretations are obtained by the weighted sum of features in the last convolutional layer.
Grad-CAM is only applied to the last convolutional layer in \cite{8237336}.
Meanwhile, the channel weight in Grad-CAM is obtained from an individual image.
Fig. \ref{fig:5layers_complex} shows that interpretations of Grad-CAM in former layers are not reasonable, which always concentrate on the background.
For example, for the top image in Fig. \ref{fig:5layers_complex}, Grad-CAM highlights the background but rather zebra from the pool1 layer to the pool4 layer.

In contrast, because the proposed CHIP model is based on network perturbation, our model can give visual interpretations in different layers.
The class-discriminative importance of the CHIP model is obtained from the image dataset. Therefore, CHIP captures more reliable class-discriminative interpretations in different layers.
In Fig. \ref{fig:5layers_complex}, the CHIP model can always highlight the zebra region in these layers.

Fig. \ref{fig:5layer_simple} and Fig. \ref{fig:5layers_complex} both illustrate that the visualization in a deeper layer captures better class-discriminative representation, which means the saliency of the target object is more obvious in a deeper layer.

\subsection{Class-discriminative Importance of channels for Different Classes} \label{classes}
In the section, we compare the class-discriminative importance of channels between different classes in the interpretation of VGG16 for object classification.
Here, we select three classes as examples: bullfrog, tree frog, and zebra.
The former two classes belong to the different species of frog, while the third class barely shares similarity with frog.
Because a deep layer can capture a high-level semantic representation, we select the last convolutional layer for comparison.

Fig. \ref{fig:sparse} shows the comparison of the class-discriminative importance of channels for different classes, where the vertical axis and horizontal axis represent the channel index and the class respectively.
We note the sparsity of the class-discriminative importance of channels for different classes, since only a small subset of channels are important for each class.
It also shows that similar classes (bullfrog and tree frog) have some common important channels, while different classes (tree frog and zebra) seldom have overlapped important channels.

To further illustrate this observation, we plot two Venn diagrams comparing the number of overlapped important channels among three classes.
In Fig. \ref{fig:venn3}, the left Venn diagram shows the number of overlapped channels within the top 10 important channels.
The right one shows the number of overlapped channels in the important channels whose importance are larger than a thousandth of the highest one for each class. 
In the right Venn diagram, the number of important channels for each class is not uniform.
Fig. \ref{fig:venn3} illustrates that the number of overlapped important channels between similar classes (bullfrog and tree frog) is larger than that between dissimilar classes (tree frog and zebra).

\section{Conclusion} 
\label{sec:conclusion}
In the work, we proposed a novel CHIP model, which can provide visual interpretations for the predictions of networks without requiring re-training.
Further, we combine visual interpretations in the first and the last convolutional layers to obtain Refined CHIP visual interpretation that is both class-discriminative and high-resolution.
Through experiment evaluation, we have demonstrated that the proposed interpretation model can provide more reasonable visual interpretation compared with previous interpretation methods. 
The proposed CHIP model also outperforms other visual interpretation methods in weakly-supervised object localization task.

\bibliographystyle{IEEEtran}
\bibliography{CHIP_ref}

\end{document}